\pdfoutput=1

\documentclass[11pt]{article}

\usepackage[final]{acl}

\usepackage{times}
\usepackage{latexsym}

\usepackage[T1]{fontenc}

\usepackage[utf8]{inputenc}

\usepackage{microtype}

\usepackage{inconsolata}

\usepackage{graphicx}
\usepackage{todonotes}

\usepackage{tikz}
\usetikzlibrary{trees,positioning,shapes,shadows,arrows}
\tikzset{
  basic/.style  = {draw, text width=2cm, drop shadow, font=\sffamily, rectangle},
  root/.style   = {basic, rounded corners=2pt, thin, align=center, fill=white, text width=3.8cm},
  level-2/.style = {basic, rounded corners=6pt, thin,align=center, fill=white, text width=3cm},
  level-3/.style = {basic, thin, align=center, fill=white, text width=2.3cm},
  level-4/.style = {basic, thin, align=center, fill=white, text width=2.5cm}
}

\usepackage{linguex}

\title{Subjectivity in the Annotation of Bridging Anaphora}

\author{\textbf{Lauren Levine} \and 
\textbf{Amir Zeldes}  \\
  Georgetown University \\
  Department of Linguistics \\
  \texttt{\{lel76, amir.zeldes\}@georgetown.edu}}

\begin{document}
\maketitle
\begin{abstract}

Bridging refers to the associative relationship between inferable entities in a discourse and the antecedents which allow us to understand them, such as understanding what "the door" means with respect to an aforementioned "house". As identifying associative relations between entities is an inherently subjective task, it is difficult to achieve consistent agreement in the annotation of bridging anaphora and their antecedents. In this paper, we explore the subjectivity involved in the annotation of bridging instances at three levels: anaphor recognition, antecedent resolution, and bridging subtype selection. To do this, we conduct an annotation pilot on the test set of the existing GUM corpus, and propose a newly developed classification system for bridging subtypes, which we compare to previously proposed schemes. Our results suggest that some previous resources are likely to be severely under-annotated. We also find that while agreement on the bridging subtype category was moderate, annotator overlap for exhaustively identifying instances of bridging is low, and that many disagreements resulted from subjective understanding of the entities involved. 



\end{abstract}

\section{Introduction}

Bridging is an anaphoric phenomenon where a newly introduced discourse entity is dependent on an associated, non-identical antecedent entity for interpretation. The term ``bridging'' refers to a discourse participant's construction of an implicature from the entity they are currently processing back to an antecedent entity \citep{clark-1975-bridging}. This associative relation can be triggered by a broad variety of linguistic mechanisms, including lexical part-whole relations (\textit{a house - the door}) and implicit arguments (\textit{a murder - the victim}). Since the phenomenon was first commented on by \citet{clark-1975-bridging}, it has received a variety of theoretical treatments, including \citet{prince1981toward}'s closely related notion of \textit{Inferrables} which centers information status as the key component in identifying anaphoric bridging relations. Such theoretical divides have resulted in a number of different annotation formalisms varying in their definitions of bridging, as well as in their delineations of sub-varieties of bridging \citep{kobayashi-ng-2020-bridging}. While there has recently been some effort to harmonize bridging annotations across different corpora \citep{levine-zeldes-2024-unifying}, the current landscape of bridging resources remains heterogeneous. The lack of consistency in and across bridging resources largely stems from their differing definitions for bridging, as well as the subjective annotator judgments that go into identifying instances of bridging.



In this paper, we explore subjectivity in the annotation of bridging anaphora in order to understand how to account for that subjectivity and create more consistent annotations in future efforts. We examine three stages in the annotation process where annotators must make subjective judgments: (1) recognition of the bridging anaphor, (2) resolving back to its associated antecedent, and (3) identifying the subtype category of the bridging pair. To this end, we conduct an annotation pilot on the test set of an existing English corpus, GUM (v10) \citep{Zeldes2017}. While the GUM corpus includes bridging annotations, the annotation guidelines are underspecified and do not include bridging subtype annotations. This annotation pilot is a preliminary phase in the development a new  bridging resource, GUMBridge. For this effort, we develop a new classification system for bridging subtypes organized under 3 relation types: \textsc{comparison} relations, \textsc{entity} relations, and \textsc{set} relations, as well as an additional \textsc{other} category. We also create annotation guidelines for how to identify instances of bridging anaphor-antecedent pairs and how to classify them into subtypes. 

Analyzing the results of this pilot, we find on the one hand that we are able to identify substantially more and denser attestation of bridging than suggested by several previous resources. In terms of subjectivity, we find moderate agreement for the selection of the bridging subtype category and for the selection of an antecedent for a given anaphor. However, the annotator overlap in the recognition of bridging anaphora is considerably lower, despite mostly plausible precision. We conduct a qualitative evaluation of the annotations from the pilot, and we find that subjectivity plays a role in each of the three annotator judgment stages listed above, especially for recall. We explore this role for each stage, and then give recommendations on how to structure the annotation of bridging anaphora in order to account for subjectivity in annotator judgment. 




\section{Background}
\label{sec:background}

As mentioned above, there are a number of different annotation formalisms for bridging, all with somewhat different definitions of bridging as a phenomenon. In English, the evaluation of bridging resolution systems (systems which aim to automatically identify bridging anaphora and resolve back to their associative antecedents) is commonly conducted using the following three corpora: ISNotes \citep{markert-etal-2012-collective}, BASHI \citep{rosiger-2018-bashi}, and ARRAU RST \citep{poesio-artstein-2008-anaphoric, Uryupina2019AnnotatingAB}. While ARRAU RST annotates bridging instances by identifying mention pairs that establish cohesion in text and then classifies then via a set of predefined semantic relations, ISNotes and BASHI annotate bridging anaphora based on the information status of entities, considering bridging to be a sub-variety of mediated information.




The information status (IS) of an entity refers to the extent to which the entity is accessible to the reader/hearer of a discourse \citep{nissim-etal-2004-annotation}. Generally speaking, "New" information is unrecognized by the reader/hearer, while "Given" information is recognized. "Given" entities may be recognized by the reader/hearer for various reasons: the entity may have been previously introduced in the discourse (coreference), the entity may be accessible via generics/world knowledge, or, in the case of bridging, the referent of the entity may be inferred from a previous entity in the discourse. Instances of bridging and generics/world knowledge are both considered "Accessible" in that they are recognized by the reader/hearer when they are first introduced to the discourse, but only instances of bridging depend on an associative antecedent for comprehension.


\begin{table}[h!tb]
\centering
\resizebox{\columnwidth}{!}{%
\begin{tabular}{lccc}
\hline
 & Tokens & \begin{tabular}[c]{@{}c@{}}Bridging\\ Instances\end{tabular} & \begin{tabular}[c]{@{}c@{}}Bridging per \\ 1k Tokens\end{tabular} \\ \hline
ARRAU RST & 229k & 3.7k & 16.5 \\
ISNotes & 40k & 663 & 16.6 \\
BASHI & 58k & 459 & 7.9 \\
GUM (v10; full) & 228k & 1.9k & 8.3 \\
GUM (v10; test only) & 26k & 222 & 8.5 \\
GUMBridge (v0.1) & 26k & 401 & 15.4 \\ \hline
\end{tabular}%
}
\caption{Frequency of bridging instances several English bridging resources.}
\label{tab:corpora_stats}
\end{table}

There are also a number of other existing bridging resources: in English, GUM, SciCorp \cite{roesiger-2016-scicorp}, corefpro \cite{grishina-2016-experiments}, RED (Richer Event Descriptions, \citealt{ogorman-etal-2016-richer}); as well as in other languages: GRAIN \citep{schweitzer-etal-2018-german} and DIRNDL \citep{Eckart2012} in German, PDT \citep{nedoluzhko-etal-2009-coding} in Czech, and PCC \citep{ogrodniczuk-zawislawska-2016-bridging} in Polish, to name a few. There have additional been efforts in areas closely related to bridging, such as \citet{recasens-etal-2010-typology}, which puts forward a typology for classifying near-identity relations (NIDENT) for coreference, and \citet{modjeska2004resolving}'s work on other-anaphora, which we now consider a subtype of bridging. We provide background on ISNotes, BASHI, and ARRAU RST, as they are commonly used in bridging resolution evaluation \cite{yu-etal-2022-codi, kobayashi-etal-2023-pairspanbert}, and they illustrate diverging perspectives on identifying bridging instances. Table \ref{tab:corpora_stats} shows comparative statistics for these three resources, the original GUM bridging annotations, and the bridging annotations produced in the GUMBridge annotation pilot described in this paper.


ISNotes is a corpus of 50 Wall Street Journal (WSJ) documents from the OntoNotes corpus \citep{weischedel2011ontonotes} annotated for fine-grained information status. ISNotes distinguishes three main categories of IS: \texttt{New}, \texttt{Old}, and \texttt{Mediated}. \texttt{Old} information is that which known to the hearer and/or has been refereed to previously, while \texttt{New} information is introduced for the first time. \texttt{Mediated} information has not been introduced before, but is not independently comprehensible, requiring either an inference from a previous mention or from general/real-world knowledge. Within the \texttt{Mediated} category, there are six subcategories, including \texttt{bridging}. The corpus contains 663 instances of bridging in the \texttt{mediated/bridging} category, and there are an additional 253 instances of comparative anaphora in the \texttt{mediated/comparison} category, which is considered a variety of bridging (\textasciitilde 16.6 bridging instances per 1k tokens). \citet{markert-etal-2012-collective} report Cohen's $\kappa$ for annotator pairs, ranging \textasciitilde0.6-0.7 for \texttt{mediated/bridging}, and \textasciitilde0.8 for \texttt{mediated/comparison}. They note that the agreement for \texttt{mediated/bridging} is more annotator dependent relative to the other IS categories. 

The BASHI corpus is also annotated on top of 50 WSJ documents from the OntoNotes corpus, and it includes a total of 459 bridging pairs (\textasciitilde 7.9 bridging instances per 1k tokens). \citet{rosiger-2018-bashi} introduces the contrast between referential bridging and lexical bridging, where referential bridging is a properly anaphoric relation (antecedent is required for the interpretation of the anaphor) and lexical bridging is a non-anaphoric semantic relation between two entities. The corpus specifically contains annotations only for referential bridging, not lexical bridging. The bridging instances in BASHI have the subtypes definite, indefinite, and comparative anaphora. Annotator agreement is reported for these categories individually and together. The joint agreement for identifying bridging pairs is 59.3\%, with a higher rate for comparative anaphora at 71.4\% and lower agreement for definite at 63.8\% and indefinite at 42.3\%.


ARRAU is a multi-genre corpus covering a variety of anaphoric phenomena, composed of 4 sub-corpora, each with its own annotation specifications. ARRAU RST is the largest sub-corpus, and also the one most used in evaluation for bridging resolution. It is composed of WSJ news data, and it includes 3,777 bridging annotations (\textasciitilde 16.5 bridging instances per 1k tokens). ARRAU's bridging annotation connects related mentions which establish "entity coherence" via non-identity relations, but as this casts a very broad scope, annotation is limited to a fixed set of semantic relations. The corpus uses an inventory of 9 bridging subtypes for annotation: \texttt{possession}, \texttt{element-set}, \texttt{subset-set}, anaphora marked with `other', along with accompanying inverse relations of the previous, and an additional \texttt{under-specified} relation. The annotation schema and guidelines for bridging in ARRAU were extended from the GNOME project \citep{poesio-2004-discourse}. Coders in the GNOME project displayed high agreement (95.2\%) in the choice of bridging subtype labels from its fixed set of relations, but low recall (22\%) in unanimously identifying instances of bridging.

Limiting annotation to a predefined set of relations restricts the scope of bridging as a phenomenon, but also aims to increase consistency in the annotation. However, as has been noted in \citet{rosiger-2018-bashi}, annotating from predefined relations can also introduce false positives, in the case that an instance of a semantic relation is not actually a case of associative anaphoric reference that would constitute referential bridging. For instance, the case of \textit{Europe - Spain} displays a meronomy relation, but it is not anaphoric because \textit{Spain} can be interpreted without reference to \textit{Europe}. Annotating from an information status informed perspective aims to avoid such false positives, providing a more concrete linguistic criteria for identifying instances of bridging when compared to the notion of "entity coherence", and eliminating the need to only annotate a predefined set of relations for scoping reasons. However, this information status based approach also greatly widens the scope of what should be considered bridging, which in turn increases the influence of subjective judgment by annotators. As such, in order to forward an information status informed annotation perspective, we must develop means of dealing with additional subjectivity it produces.

As we can see in Table \ref{tab:corpora_stats}, there has been considerable variation in the frequency of bridging annotations in previous resources, with ARRAU RST (counting both lexical and referential bridging) and ISNotes identifying bridging instances with approximately twice the rate per 1k tokens as the annotations in BASHI and GUM v10. This suggests that some previous bridging resources, such as BASHI and GUM, have likely been under-annotated for bridging instances and prompts a need for the reexamination of bridging annotation procedures. 




\section{Annotation Pilot}

The analysis on subjectivity in the annotation of bridging instances in this paper is conducted using the results of an annotation pilot for the creation of a new bridging resource called GUMBridge. Built on top of GUM, an existing multi-genre corpus of English, GUMBridge aims to unite aspects of currently existing formalisms: using an information status-informed view of identifying bridging instances (as in ISNotes and BASHI), followed by subtype categorization using a taxonomy of semantic relations (as in ARRAU). Additionally, GUMBridge aims to add genre diversity to the core English bridging resources, as ISNotes, BASHI, and ARRAU RST are all composed of WSJ news data from more than 30 years ago, offering little to analyze in terms on genre diversity. While the development of this resource is still underway, an adjudicated version of the bridging annotations for the GUMBridge test set (version 0.1) is released with this paper\footnote{https://github.com/lauren-lizzy-levine/gumbridge}. The details of this adjudication process are described in Section \ref{sec:adjudication}. The guidelines for identifying instances of bridging (v0.1) are described in Section \ref{sec:identify_bridge}, and the classification system for bridging subtypes (v0.1) is described in Section \ref{sec:taxonomy}.



\subsection{Identifying Bridging Instances}
\label{sec:identify_bridge}

In the GUMBridge annotation effort, we adopt an information status-informed perspective on identifying instances of bridging anaphora. As stated in Section \ref{sec:background}, the information status of an entity refers to the extent to which an entity is accessible to the reader/hearer of a discourse upon its introduction. We say that an entity is ``Accessible'' if it has not been mentioned before but its reference is inferable for a reader/header. Bridging occurs when the first mention of an entity is ``Accessible'' via an inference from a previous, non-identical entity in the discourse. In contrast with entities which are accessible due to being generic, or being part of world knowledge or the discourse situation, the bridging anaphor is not accessible by itself, but dependent on the previous entity for interpretation. Annotators are provided with an overview of this definition of bridging and accessibility and are instructed to consider the following when deciding whether a particular entity is a bridging anaphor:


\begin{enumerate}
    \item Do you judge this entity to be to some degree accessible in the discourse?
    \item Does that accessibility rely on the understanding of a previous entity in the discourse? If so, identify that previous entity’s most recent mention.
\end{enumerate}

If the entity passes the above criteria, it is a bridging anaphor and the previous entity is its associative antecedent. Once identified, a bridging pair can then be assigned a subtype category as described in the following section.


\subsection{Classification of Bridging Subtypes}
\label{sec:taxonomy}

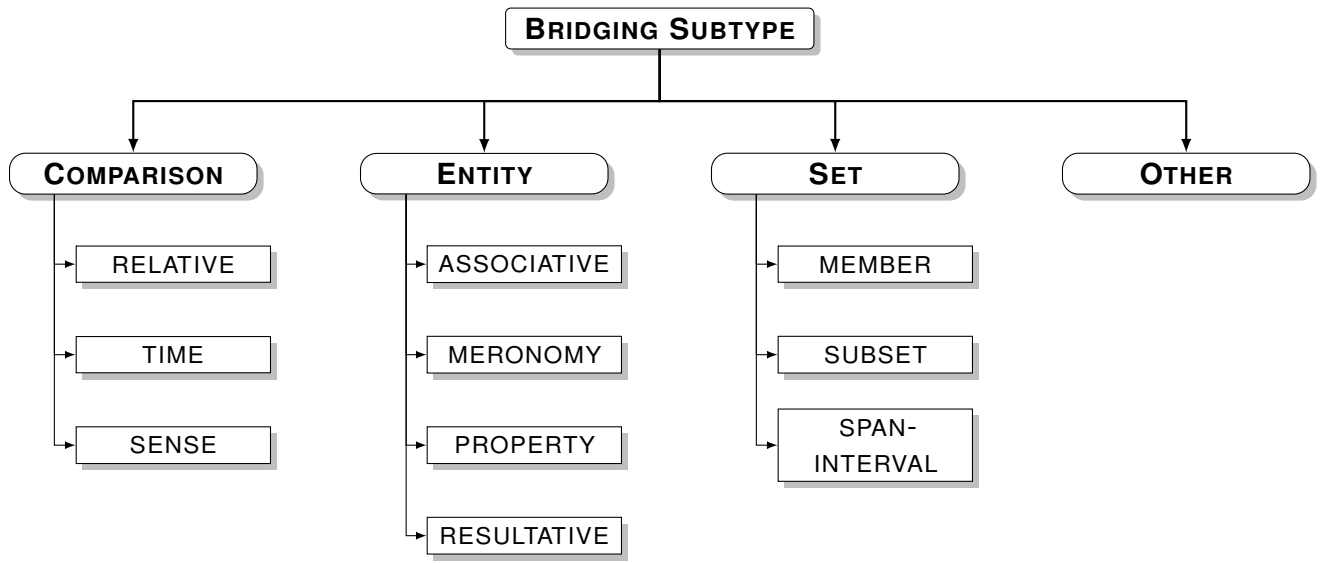
\begin{figure*}
    \centering
\begin{tikzpicture}[
  level 1/.style={sibling distance=12em, level distance=5em},
   {edge from parent fork down},
  edge from parent/.style={->,solid,black,thick,draw}, 
  edge from parent path={(\tikzparentnode.south) -- (\tikzchildnode.north)},
  >=latex, node distance=1.2cm, edge from parent fork down]

\node[root] (c0) {\textbf{\textsc{Bridging Subtype}}}

  child {node[level-2] (c1) {\textbf{\textsc{Comparison}}}}
  child {node[level-2] (c2) {\textbf{\textsc{Entity}}}}
  child {node[level-2] (c3) {\textbf{\textsc{Set}}}}
  child {node[level-2] (c4) {\textbf{\textsc{Other}}}};

\begin{scope}[every node/.style={level-3}]
\node [below of = c1, xshift=15pt] (c11) {\textsc{relative}};
\node [below of = c11] (c12) {\textsc{time}};
\node [below of = c12] (c13) {\textsc{sense}};

\node [below of = c2, xshift=15pt] (c21) {\textsc{associative}};
\node [below of = c21] (c22) {\textsc{meronomy}};
\node [below of = c22] (c23) {\textsc{property}};
\node [below of = c23] (c24) {\textsc{resultative}};

\node [below of = c3, xshift=15pt] (c31) {\textsc{member}};
\node [below of = c31] (c32) {\textsc{subset}};
\node [below of = c32] (c33) {\textsc{span-interval}};

\end{scope}


\foreach \value in {1,2,3}
  \draw[->] (c1.195) |- (c1\value.west);

\foreach \value in {1,...,4}
  \draw[->] (c2.195) |- (c2\value.west);
  
\foreach \value in {1,2,3}
  \draw[->] (c3.195) |- (c3\value.west);

  
\end{tikzpicture}
    \caption{Bridging Subtype Classification in GUMBridge v0.1.}
    \label{fig:subtype_taxonomy}
\end{figure*}

In order to categorize the varieties of bridging present in GUMBridge, we create a new classification system for bridging subtypes. The classification system is composed of 11 categories, 10 of which are organized under 3 relation types: \textsc{comparison} relations, \textsc{entity} relations, and \textsc{set} relations, and an additional \textsc{other} category. The bridging subtype classification system developed for GUMBridge (v0.1) is shown in Figure \ref{fig:subtype_taxonomy}. A brief description of each of the bridging subtypes follows below. A brief comparison to the bridging subtypes of ARRAU is included in Appendix \ref{sec:appendix_comp}.

\paragraph{\textsc{comparison-relative}} The anaphor is preceded by a comparative marker (other, another, same, more, etc.), ordinal (second, third, etc.), or comparative adjective (larger, smaller, etc.), which implies a comparison to the antecedent (or vice versa). 

\ex. \underline{Several women} walked into the room. \textbf{Other women} soon followed.

\paragraph{\textsc{comparison-time}} The anaphor refers to a specific time/time frame which is understandable with reference to the time/time frame expressed by the antecedent (or vice versa). 

\ex. I went shopping \underline{Wednesday, March 3rd}. I will go again \textbf{the following Wednesday}.




\paragraph{\textsc{comparison-sense}} The type of the anaphor is omitted but inferable via comparison to the antecedent (or vice versa).


\ex. I’ve been to the \underline{Chinese restaurant}. I want to go to \textbf{the Italian one}.

\paragraph{\textsc{entity-associative}} The anaphor is an attribute or closely associated entity of the antecedent (or vice versa). This frequently manifests as implicit arguments of a predicate as in example \ref{ex:arg}, relational nouns as in example \ref{ex:relational}, and prototypical associations as in example \ref{ex:prototype}: 

\ex. There was \underline{a murder} last night. \textbf{The victim} has yet to be identified.
\label{ex:arg}

\ex. There is \underline{a child} in the park. \textbf{The parent} must be nearby.
\label{ex:relational}

\ex. I went to \underline{a wedding} last week. \textbf{The reception} was really fun.
\label{ex:prototype}

\paragraph{\textsc{entity-meronomy}} The anaphor is a subunit of the antecedent (or vice versa), i.e., there is some part-whole relation between the anaphor and the antecedent.

\ex. I saw \underline{a large house} by the lake. \textbf{The door} was red. 

\paragraph{\textsc{entity-property}} The anaphor is a physical or intangible property of the antecedent (or vice versa). For example: smell, length, style, etc.

\ex. I picked up \underline{a bouquet of roses}. \textbf{The scent} was lovely. 

\paragraph{\textsc{entity-resultative}} The anaphor is logically inferable from the antecedent (or vice versa). This is typically the result of a transformative or product producing process, such as cooking.\footnote{This subtype subsumes the \textsc{transformed} type proposed by \citet{fang-etal-2022-take} specifically for recipe outcomes.}

\ex. Though \underline{my flour} was a strange texture, \textbf{the bread} came out perfectly.


\paragraph{\textsc{set-member}} The anaphor is an element of the antecedent set (or vice versa). 

\ex. I got \underline{several books} for my birthday. \textbf{The mystery novel} was my favorite. 

\paragraph{\textsc{set-subset}} The anaphor is a subset of the antecedent set (or vice versa).

\ex. \underline{A group of students} entered the hall. \textbf{The boys} wore neckties with their uniforms. 

\paragraph{\textsc{set-span-interval}} The anaphor is a sub-span of the spatial or temporal antecedent interval (or vice versa). 

\ex. If you want to meet up on \underline{Sunday}, I will be free in \textbf{the morning}.

\paragraph{\textsc{other}} The anaphor and antecedent fit the criteria for identifying a bridging pair, but do not fall into any of the bridging subtypes detailed above. For instance, \citet{ogrodniczuk-zawislawska-2016-bridging} give examples of metareference:

\ex. I went to \underline{Sensational Cakes} yesterday, but I didn't think \textbf{the cakes} were very good.
\label{ex:meta}


Metareference allows for reference back to a name or label, as in example \ref{ex:meta}. 
Such instances are unique and interesting enough to wish not to shoehorn them into another category, but are not common enough to warrant a separate category in the subtype classification.

As stated in Section \ref{sec:identify_bridge}, the criterion for identifying instances of bridging is anaphoric, relying on information status and resolution back to an associative antecedent. The subtype labels primarily allow us to understand how the phenomenon manifests in a discourse, and, as such, there is no theoretical reason to limit the number of subtypes that can apply to an instance of bridging to just one. Indeed, there are cases of bridging where multiple subtypes may apply:

\ex. \underline{Several women} walked into the room. \textbf{One} left immediately.
\label{ex:sense_member}

\ex. I will come to visit \underline{this week}, as I could not come \textbf{the previous week}.
\label{ex:relative_time}

Example \ref{ex:sense_member} shows an instance for which \textsc{comparison-sense} and \textsc{set-member} both apply, while example \ref{ex:relative_time} show a case where \textsc{comparison-relative} and \textsc{comparison-time} apply. In this annotation pilot, annotators where instructed to select a single bridging subtype, prioritizing certain categories over others if they occurred together. However, in principle, all applicable subtypes could be annotated. In our subsequent efforts to annotate the remaining data in GUM and produce a full version of GUMBridge, we intend to support the annotation of multiple bridging subtypes for a single bridging pair for the entire corpus. 

\subsection{Annotation Procedure}

The GUMBridge annotation pilot was conducted on the test set of the existing GUM (v10) corpus, which consists of 26 documents (\textasciitilde26k tokens) across 16 genres (academic writing, biographies, courtroom transcripts, essays, fiction, how-to guides, interviews, letters, news, online forum discussions, podcasts, political speeches, spontaneous face to face conversations, textbooks, travel guides, and vlogs). The GUM corpus already includes annotations for entity spans, coreference,\footnote{The coreference scheme considers all mentions eligible for bridging, including indefinite anaphors, discourse deixis to non-nominal antecedents and more, see \citet{Zeldes2022} for a detailed discussion.} and information status, i.e., "New", "Given", and "Accessible" (not including accessibility from instances of bridging).

The documents of the test set were double annotated, with one author of this paper acting as Annotator A and various linguistics graduate students acting as Annotator B for different documents in the test set. Each of the 8 annotators acting as Annotator B was assigned between 2 and 4 documents of the test set. The annotation was completed using the GitDox annotation interface \citep{Zhang2017GitDOXAL}. For the existing entity annotations in the document, the annotator was instructed to identify whether the entity is a bridging anaphor, and, if so, create a link between the anaphor and its associative antecedent. The annotator was instructed to also update the IS of the bridging anaphor to ``Accessible'' and select a bridging subtype annotation for the anaphor. The full annotation guidelines provided to the annotators are included as supplementary materials. 



\subsection{Agreement Study}

In Table \ref{tab:agree}, we provide agreement numbers for three stages of the bridging annotation process: anaphor recognition, antecedent resolution, and subtype categorization. 

\begin{table}[h!tb]
\centering
\resizebox{\columnwidth}{!}{%
\begin{tabular}{cccc}
\hline
 & \textbf{Precision} & \textbf{Recall} & \textbf{F1 Score} \\ \hline
\begin{tabular}[c]{@{}c@{}}Anaphor \\ Recognition\end{tabular} & 0.44 & 0.34 & 0.38 \\ \cline{1-1}
\begin{tabular}[c]{@{}c@{}}Anaphor+Antecedent\\ Recognition\end{tabular} & 0.32 & 0.25 & 0.28 \\ \hline
 & & \textbf{Accuracy} & \\
 \hline
 \begin{tabular}[c]{@{}c@{}}Antecedent\\ Resolution\end{tabular} & & 0.72 & \\
\hline
 & & \textbf{Cohen's $\kappa$} & \\
 \hline
 \begin{tabular}[c]{@{}c@{}}Bridging\\ Subtype\end{tabular} & & 0.58 & \\
 \hline
\end{tabular}%
}
\caption{GUMBridge pilot inter-annotator agreement.}
\label{tab:agree}
\end{table}

For the recognition of bridging pairs (anaphor+antecedent) and recognition of the bridging anaphor alone, we give the PRF of Annotator B relative to Annotator A. We see that the F1 for bridging anaphor recognition is 0.38, and the F1 for bridging pair recognition is only 0.28. As the recognition of bridging pairs is inherently limited by the recognition of the anaphor, we also give the accuracy of Annotator B selecting the antecedent entity when both annotators agree on the bridging anaphor, which is 72\% of a total of 133 cases. Finally, for the 96 instances where both annotators agreed on the anaphor and antecedent of a bridging pair, the Cohen's Kappa for the bridging subtype annotation is 0.58, which indicates moderate agreement. These numbers suggest that the key hurdle is in anaphor recognition, though antecedent resolution and subtype labeling are also non-trivial.

\begin{figure}
  \centering
  \includegraphics[width=1\linewidth]{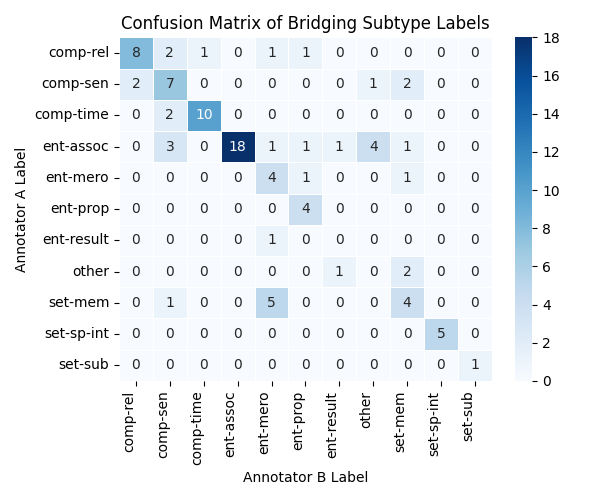}
  \caption{Confusion matrix of bridging subtypes for bridging instances with matching anaphor and antecedent annotations.}
  \label{fig:bridgetype_cm}
\end{figure}

In Figure \ref{fig:bridgetype_cm}, we show a confusion matrix of the bridging subtype labels assigned by Annotator A and Annotator B to the overlapping bridging pairs. We see that the subtypes with the most overlap are the \textsc{comparison} categories and \textsc{entity-associative}. And while there are some categories for which the disagreement is spread among a number of categories, we see that the categories of \textsc{entity-meronomy} and \textsc{set-member} are particularly confusable, which indicates how part-whole and set-member relations can be quite similar. The categories of \textsc{entity-associative} and \textsc{other} are also particularly confusable, which speaks to how \textsc{entity-associative} may be an overly broad category. Although agreement on bridging subtype annotation is moderate, it is clear that refinement in the guidelines for the categories is still needed. However, as agreement on the identification of bridging instances is substantially lower, recognition of bridging anaphora forms the limiting point in the annotation process.




\subsection{Data Adjudication}
\label{sec:adjudication}

As shown in the previous section, the results of the annotation pilot had low annotator agreement, necessitating a qualitative analysis of annotations to determine the cause of the disagreements. As a part of this process, the annotations from the pilot were adjudicated to produce a single set of reference bridging annotations for the test set of GUMBridge (v0.1), available with the release of this paper under the Creative Commons Attribution (CC-BY) version 4.0 license. The composition of the GUMBridge test set by bridging subtype after the adjudication is shown in Appendix \ref{sec:appendix_counts}. The test set of GUMBridge has a total of 401 bridging annotations, with an average of 15.4 bridging instances per 1k tokens. This is on par with the higher rate of bridging instances per 1k tokens found in ISNotes and ARRAU RST as shown in in Table \ref{tab:corpora_stats}. While the limited size of the data set annotated in this pilot limits our ability to make observations on genre effects, for completeness, a breakdown of the bridging relation types observed in each genre is included in Appendix \ref{sec:appendix_genre} . 

Notably, the number of instances in the test set of the GUM (v10) annotations nearly doubles, going from 222 instances of bridging to 401 in GUMBridge test, suggesting a significant improvement in coverage of bridging instances in this new annotation effort. Even though there is less consistency in this annotation effort compared to some of those discussed in Section \ref{sec:background}, numbers suggest higher recall, which allows us to capture a greater scope of bridging instances. As bridging is generally a sparse phenomenon, the annotations can be manually reviewed and validated in the adjudication process even if initial agreement is low. As such, we believe it is preferable to favor a high recall method of annotation and eliminate false positives upon review, rather than risk many interesting cases that will remain unidentified. 

The adjudication process involved comparing all of the diverging judgments from Annotator A and Annotator B  at the level of anaphor, antecedent, and subtype. Table \ref{tab:disagreement_types} shows the proportion of such disagreements in the pilot annotations. Of the 172 instances that Annotator B labeled as bridging which Annotator A initially did not label as bridging at all, upon reevaluation, it was concluded that 64 (37\%) could reasonably be considered a form of bridging. Many of these judgments relied on subjective understanding of the discourse entities involved. In the following section, we provide an analysis of the impact of subjectivity in this annotation pilot and how it may be better handled in the future.

\begin{table}[]
\centering
\resizebox{0.6\columnwidth}{!}{%
\begin{tabular}{ll}
\hline
Completely Matching & 61 \\
Different Subtype & 35 \\
Different Antecedent & 37 \\
Annotator B Only & 172 \\
Annotator A Only & 257 \\ \hline
\textbf{Total} & 562 \\ \hline
\end{tabular}%
}
\caption{Counts of annotator agreement/disagreement types in GUMBridge pilot annotations.}
\label{tab:disagreement_types}
\end{table}

\section{Subjectivity in Bridging Annotation}

Previous work on subjectivity in the development of linguistic data has heavily featured areas where annotator judgments can be highly variable, such as hate speech detection and sentiment analysis (e.g., \citet{waseem-2016-racist, kenyon-dean-etal-2018-sentiment}), though attention has also been given to tasks which seem more objective, such as part of speech annotation (e.g., \citet{plank-etal-2014-linguistically}). Several works discuss the paradigms for and implications of including subjective judgments in annotation efforts, rather than trying to eliminate all ambiguity \citep{ovesdotter-alm-2011-subjective, rottger-etal-2022-two}. Ultimately, the appropriate approach depends on the linguistic task at hand and what the researchers are hoping to achieve with the annotation effort. 

Although detailed guidelines are provided to annotators in this paper's annotation pilot, subjective judgment is still an inherent part of the annotation of bridging instances, as annotators are making decisions based off their understanding of the implicit relationships that exist between entities in a discourse. As previously noted, there are three decision points in the annotating of bridging instances that can introduce subjective judgment: (1) recognition of the bridging anaphor, (2) identifying the  corresponding associative antecedent, and (3) selecting the bridging subtype category of the pair. The sections below give examples to illustrate the unique considerations regarding subjectivity that are present at each of these annotation stages.

\subsection{Subtype Categorization}

Selecting a bridging subtype category relies on understanding the relationship between the anaphor and the antecedent in a bridging pair. The exact nature of the relationship between two entities is dependent on the annotator's subjective conception of the two entities. It is possible that a lack of familiarity with related entities may cause annotation errors:

\ex. \underline{the cuttings} $\rightarrow$ \textbf{the first pad}
\label{ex:cutting}

In example \ref{ex:cutting}, ``the cuttings'' refer to cactus cuttings, each of which is a whole pad. Without this particular knowledge, it would be reasonable for an annotator to assume that a pad is a portion of a cutting or that a cutting is a portion of a pad. 

There may be additional uncertainty in interpreting an entity based on the context of the discourse:

\ex. \underline{peppermint plants} $\rightarrow$ \textbf{the mint}
\label{ex:mint}

In the discourse context of example \ref{ex:mint}, it is unclear whether ``the mint'' is referring back to a specific part of the peppermint plant (e.g.~the leaves), or whether it is an instance of synecdoche, referring to the plant as a whole.

There are also instances where multiple subtypes are possible in the context of the discourse:

\ex. \underline{some basil} $\rightarrow$ \textbf{seed}
\label{ex:basil}

In the discourse context of example \ref{ex:basil}, a question is being posed whether ``some basil'' can be grown from ``seed''. As such, it is reasonable to say that the basil comes from the seed in which case the subtype would be \textsc{entity-resultative}. However, it is also reasonable to say that seed is a part of the basil plant, in which case the subtype would be \textsc{entity-meronomy}. In such cases, it is necessary to have a priority hierarchy for deciding which bridging subtype category should be assigned, or we must allow for multiple subtype annotations. In future work, we intend to support the annotation of multiple bridging subtypes for the entire GUMBridge corpus.

\subsection{Antecedent Selection}

When an annotator is selecting the associative antecedent of a bridging anaphor, there are also opportunities for subjective judgments to be made. In some cases, it is possible that multiple preceding entities could be reasonable candidates for a bridging antecedent:

\ex. \underline{your mouth} $\rightarrow$ \textbf{other body parts…} \\ \underline{teeth} $\rightarrow$ \textbf{other body parts…}
\label{ex:body_parts}

The example \ref{ex:body_parts} refers to a case where a dental cast is being made and the narrator wonders what other body parts can be given the same treatment. It is not clear whether ``the other body parts'' are more appropriately in contrast with the ``mouth'' or ``teeth'', or even both, if we accept both teeth and mouths as body parts.  

There is also the possibility for disagreement on the denotation of the anaphor:

\ex. \underline{the bridge} $\rightarrow$ \textbf{the edge} \\ \underline{the upper levels} $\rightarrow$ \textbf{the edge}
\label{ex:edge}

In example \ref{ex:edge}, the narrator considers looking over ``the edge'', and it is unclear whether it is the edge of a particular bridge, or if it is the edge of some general upper level. In such cases, it may be beneficial to impose an easy to execute heuristic, such as selecting the option nearer to the bridging anaphor, assuming we are aiming for a single reference decision. Note that this is different from cases in which multiple labels apply, since the two interpretations, while both possible, are mutually exclusive.


\subsection{Anaphor Identification}

When identifying a bridging anaphor, annotators must make subjective judgments on whether an entity is accessible due to world knowledge (and hence not bridging) or whether the accessibility can be attributed to an antecedent entity. For instance, one annotator had ``Leucippus and Democritus'' bridge from ``ancient Greek philosophers'', but not ``Aristotle'' who is more widely known. This illustrates how an annotator's world knowledge may influence what they consider to be ``Accessible'' in a manner that is undesirable as it will lead to inconsistencies among annotators.  We recommend that concrete criteria for generic/world knowledge accessibility should be tied to a knowledge base, such as Wikipedia, rather than left up to individual annotator judgment. For named entities, this type of linking or Wikification is already available for GUM \cite{lin-zeldes-2021-wikigum} and will be integrated in future annotation efforts.





\section{Conclusion}

In this paper, we examine the influence of subjectivity in annotator judgment on the various stages of annotating instances of bridging. We make this examination using the resulting annotations from a pilot to create a new resource for bridging annotations, GUMBridge. We also release an adjudicated version of the bridging annotations for the preliminary test set of GUMBridge (v0.1). In subsequent work, we plan to refine the guidelines and annotation procedure used in this pilot, which we will then use to annotate the remainder of the GUM corpus (dev and train) to produce a full version of GUMBridge, as well as extending our annotations to GUM's out-of-domain challenge test set, GENTLE (GEnre Tests for Linguistic Evaluation, \citealt{aoyama-etal-2023-gentle}). As the time and effort required to manually annotate bridging limits the scalability of the annotation process, we will also investigate incorporating semi-automated methods, such as combining LLMs or other systems for bridging resolution with human correction in order to improve the efficiency of the process. 

In our development of GUMBridge test (v0.1), we found that annotators' agreement on selecting the subtype of a bridging pair was moderate, but that it was more difficult to get the annotators to align on the identification of bridging anaphora. This indicates that recognition of bridging anaphora is the stage in the annotation process that is most vulnerable to the subjective judgment of annotators, and that should be given the most consideration when trying to account for annotator subjectivity. While some subjectivity arises from the inherent ambiguity of language in context, other aspects of subjectivity can be accounted for by providing guidelines on how to decide on preferable judgments when multiple options are available.

\section*{Limitations}

The analysis presented in this paper on subjectivity in the annotation of bridging anaphora is based on a pilot annotation study for a new resource that is still in development. This limits the amount of data available for analysis to a test set of 26k tokens. The reliability of the annotation schema is also a limitation, as the results of the annotation pilot showed agreement on identification of bridging anaphora to be undesirably low, and the annotation schema/instructions will need to undergo revision in future work.


\bibliography{custom}

\begin{thebibliography}{33}
\providecommand{\natexlab}[1]{#1}

\bibitem[{Aoyama et~al.(2023)Aoyama, Behzad, Gessler, Levine, Lin, Liu, Peng,
  Zhu, and Zeldes}]{aoyama-etal-2023-gentle}
Tatsuya Aoyama, Shabnam Behzad, Luke Gessler, Lauren Levine, Jessica Lin,
  Yang~Janet Liu, Siyao Peng, Yilun Zhu, and Amir Zeldes. 2023.
\newblock \href {https://doi.org/10.18653/v1/2023.law-1.17} {{GENTLE}: A
  genre-diverse multilayer challenge set for {E}nglish {NLP} and linguistic
  evaluation}.
\newblock In \emph{Proceedings of the 17th Linguistic Annotation Workshop
  (LAW-XVII)}, pages 166--178, Toronto, Canada. Association for Computational
  Linguistics.

\bibitem[{Clark(1975)}]{clark-1975-bridging}
Herbert~H. Clark. 1975.
\newblock \href {https://aclanthology.org/T75-2034/} {Bridging}.
\newblock In \emph{Theoretical Issues in Natural Language Processing}.

\bibitem[{Eckart et~al.(2012)Eckart, Riester, and Schweitzer}]{Eckart2012}
Kerstin Eckart, Arndt Riester, and Katrin Schweitzer. 2012.
\newblock \href {https://doi.org/10.1007/978-3-642-28249-2_7} {\emph{A
  Discourse Information Radio News Database for Linguistic Analysis}}, pages
  65--76.
\newblock Springer Berlin Heidelberg, Berlin, Heidelberg.

\bibitem[{Fang et~al.(2022)Fang, Baldwin, and Verspoor}]{fang-etal-2022-take}
Biaoyan Fang, Timothy Baldwin, and Karin Verspoor. 2022.
\newblock \href {https://doi.org/10.18653/v1/2022.findings-acl.275} {What does
  it take to bake a cake? the {R}ecipe{R}ef corpus and anaphora resolution in
  procedural text}.
\newblock In \emph{Findings of the Association for Computational Linguistics:
  ACL 2022}, pages 3481--3495, Dublin, Ireland. Association for Computational
  Linguistics.

\bibitem[{Grishina(2016)}]{grishina-2016-experiments}
Yulia Grishina. 2016.
\newblock \href {https://doi.org/10.18653/v1/W16-0702} {Experiments on bridging
  across languages and genres}.
\newblock In \emph{Proceedings of the Workshop on Coreference Resolution Beyond
  {O}nto{N}otes ({CORBON} 2016)}, pages 7--15, San Diego, California.
  Association for Computational Linguistics.

\bibitem[{Kenyon-Dean et~al.(2018)Kenyon-Dean, Ahmed, Fujimoto,
  Georges-Filteau, Glasz, Kaur, Lalande, Bhanderi, Belfer, Kanagasabai,
  Sarrazingendron, Verma, and Ruths}]{kenyon-dean-etal-2018-sentiment}
Kian Kenyon-Dean, Eisha Ahmed, Scott Fujimoto, Jeremy Georges-Filteau,
  Christopher Glasz, Barleen Kaur, Auguste Lalande, Shruti Bhanderi, Robert
  Belfer, Nirmal Kanagasabai, Roman Sarrazingendron, Rohit Verma, and Derek
  Ruths. 2018.
\newblock \href {https://doi.org/10.18653/v1/N18-1171} {Sentiment analysis:
  It`s complicated!}
\newblock In \emph{Proceedings of the 2018 Conference of the North {A}merican
  Chapter of the Association for Computational Linguistics: Human Language
  Technologies, Volume 1 (Long Papers)}, pages 1886--1895, New Orleans,
  Louisiana. Association for Computational Linguistics.

\bibitem[{Kobayashi et~al.(2023)Kobayashi, Hou, and
  Ng}]{kobayashi-etal-2023-pairspanbert}
Hideo Kobayashi, Yufang Hou, and Vincent Ng. 2023.
\newblock \href {https://doi.org/10.18653/v1/2023.acl-long.383}
  {{P}air{S}pan{BERT}: An enhanced language model for bridging resolution}.
\newblock In \emph{Proceedings of the 61st Annual Meeting of the Association
  for Computational Linguistics (Volume 1: Long Papers)}, pages 6931--6946,
  Toronto, Canada. Association for Computational Linguistics.

\bibitem[{Kobayashi and Ng(2020)}]{kobayashi-ng-2020-bridging}
Hideo Kobayashi and Vincent Ng. 2020.
\newblock \href {https://doi.org/10.18653/v1/2020.coling-main.331} {Bridging
  resolution: A survey of the state of the art}.
\newblock In \emph{Proceedings of the 28th International Conference on
  Computational Linguistics}, pages 3708--3721, Barcelona, Spain (Online).
  International Committee on Computational Linguistics.

\bibitem[{Levine and Zeldes(2024)}]{levine-zeldes-2024-unifying}
Lauren Levine and Amir Zeldes. 2024.
\newblock \href {https://doi.org/10.18653/v1/2024.crac-1.5} {Unifying the scope
  of bridging anaphora types in {E}nglish: Bridging annotations in {ARRAU} and
  {GUM}}.
\newblock In \emph{Proceedings of The Seventh Workshop on Computational Models
  of Reference, Anaphora and Coreference}, pages 41--51, Miami. Association for
  Computational Linguistics.

\bibitem[{Lin and Zeldes(2021)}]{lin-zeldes-2021-wikigum}
Jessica Lin and Amir Zeldes. 2021.
\newblock \href {https://doi.org/10.18653/v1/2021.law-1.18} {{W}iki{GUM}:
  Exhaustive entity linking for wikification in 12 genres}.
\newblock In \emph{Proceedings of the Joint 15th Linguistic Annotation Workshop
  (LAW) and 3rd Designing Meaning Representations (DMR) Workshop}, pages
  170--175, Punta Cana, Dominican Republic. Association for Computational
  Linguistics.

\bibitem[{Markert et~al.(2012)Markert, Hou, and
  Strube}]{markert-etal-2012-collective}
Katja Markert, Yufang Hou, and Michael Strube. 2012.
\newblock \href {https://aclanthology.org/P12-1084/} {Collective classification
  for fine-grained information status}.
\newblock In \emph{Proceedings of the 50th Annual Meeting of the Association
  for Computational Linguistics (Volume 1: Long Papers)}, pages 795--804, Jeju
  Island, Korea. Association for Computational Linguistics.

\bibitem[{Modjeska(2004)}]{modjeska2004resolving}
Natalia~N Modjeska. 2004.
\newblock Resolving other-anaphora.

\bibitem[{Nedoluzhko et~al.(2009)Nedoluzhko, M{\'\i}rovsk{\'y}, and
  Pajas}]{nedoluzhko-etal-2009-coding}
Anna Nedoluzhko, Ji{\v{r}}{\'\i} M{\'\i}rovsk{\'y}, and Petr Pajas. 2009.
\newblock \href {https://aclanthology.org/W09-3017} {The coding scheme for
  annotating extended nominal coreference and bridging anaphora in the {P}rague
  dependency treebank}.
\newblock In \emph{Proceedings of the Third Linguistic Annotation Workshop
  ({LAW} {III})}, pages 108--111, Suntec, Singapore. Association for
  Computational Linguistics.

\bibitem[{Nissim et~al.(2004)Nissim, Dingare, Carletta, and
  Steedman}]{nissim-etal-2004-annotation}
Malvina Nissim, Shipra Dingare, Jean Carletta, and Mark Steedman. 2004.
\newblock \href {https://aclanthology.org/L04-1402/} {An annotation scheme for
  information status in dialogue}.
\newblock In \emph{Proceedings of the Fourth International Conference on
  Language Resources and Evaluation ({LREC}`04)}, Lisbon, Portugal. European
  Language Resources Association (ELRA).

\bibitem[{O{'}Gorman et~al.(2016)O{'}Gorman, Wright-Bettner, and
  Palmer}]{ogorman-etal-2016-richer}
Tim O{'}Gorman, Kristin Wright-Bettner, and Martha Palmer. 2016.
\newblock \href {https://doi.org/10.18653/v1/W16-5706} {Richer event
  description: Integrating event coreference with temporal, causal and bridging
  annotation}.
\newblock In \emph{Proceedings of the 2nd Workshop on Computing News Storylines
  ({CNS} 2016)}, pages 47--56, Austin, Texas. Association for Computational
  Linguistics.

\bibitem[{Ogrodniczuk and
  Zawis{\l}awska(2016)}]{ogrodniczuk-zawislawska-2016-bridging}
Maciej Ogrodniczuk and Magdalena Zawis{\l}awska. 2016.
\newblock \href {https://doi.org/10.18653/v1/W16-0703} {Bridging relations in
  {P}olish: Adaptation of existing typologies}.
\newblock In \emph{Proceedings of the Workshop on Coreference Resolution Beyond
  {O}nto{N}otes ({CORBON} 2016)}, pages 16--22, San Diego, California.
  Association for Computational Linguistics.

\bibitem[{Ovesdotter~Alm(2011)}]{ovesdotter-alm-2011-subjective}
Cecilia Ovesdotter~Alm. 2011.
\newblock \href {https://aclanthology.org/P11-2019/} {Subjective natural
  language problems: Motivations, applications, characterizations, and
  implications}.
\newblock In \emph{Proceedings of the 49th Annual Meeting of the Association
  for Computational Linguistics: Human Language Technologies}, pages 107--112,
  Portland, Oregon, USA. Association for Computational Linguistics.

\bibitem[{Plank et~al.(2014)Plank, Hovy, and
  S{\o}gaard}]{plank-etal-2014-linguistically}
Barbara Plank, Dirk Hovy, and Anders S{\o}gaard. 2014.
\newblock \href {https://doi.org/10.3115/v1/P14-2083} {Linguistically debatable
  or just plain wrong?}
\newblock In \emph{Proceedings of the 52nd Annual Meeting of the Association
  for Computational Linguistics (Volume 2: Short Papers)}, pages 507--511,
  Baltimore, Maryland. Association for Computational Linguistics.

\bibitem[{Poesio(2004)}]{poesio-2004-discourse}
Massimo Poesio. 2004.
\newblock \href {https://aclanthology.org/W04-0210/} {Discourse annotation and
  semantic annotation in the {GNOME} corpus}.
\newblock In \emph{Proceedings of the Workshop on Discourse Annotation}, pages
  72--79, Barcelona, Spain. Association for Computational Linguistics.

\bibitem[{Poesio and Artstein(2008)}]{poesio-artstein-2008-anaphoric}
Massimo Poesio and Ron Artstein. 2008.
\newblock \href {https://aclanthology.org/L08-1091/} {Anaphoric annotation in
  the {ARRAU} corpus}.
\newblock In \emph{Proceedings of the Sixth International Conference on
  Language Resources and Evaluation ({LREC}`08)}, Marrakech, Morocco. European
  Language Resources Association (ELRA).

\bibitem[{Prince(1981)}]{prince1981toward}
Ellen~F. Prince. 1981.
\newblock Toward a taxonomy of given-new information.
\newblock \emph{Radical pragmatics}, pages 223--255.

\bibitem[{Recasens et~al.(2010)Recasens, Hovy, and
  Mart{\'i}}]{recasens-etal-2010-typology}
Marta Recasens, Eduard Hovy, and M.~Ant{\`o}nia Mart{\'i}. 2010.
\newblock \href {https://aclanthology.org/L10-1103/} {A typology of
  near-identity relations for coreference ({NIDENT})}.
\newblock In \emph{Proceedings of the Seventh International Conference on
  Language Resources and Evaluation ({LREC}'10)}, Valletta, Malta. European
  Language Resources Association (ELRA).

\bibitem[{Roesiger(2016)}]{roesiger-2016-scicorp}
Ina Roesiger. 2016.
\newblock \href {https://aclanthology.org/L16-1275} {{S}ci{C}orp: A corpus of
  {E}nglish scientific articles annotated for information status analysis}.
\newblock In \emph{Proceedings of the Tenth International Conference on
  Language Resources and Evaluation ({LREC}'16)}, pages 1743--1749,
  Portoro{\v{z}}, Slovenia. European Language Resources Association (ELRA).

\bibitem[{R{\"o}siger(2018)}]{rosiger-2018-bashi}
Ina R{\"o}siger. 2018.
\newblock \href {https://aclanthology.org/L18-1058/} {{BASHI}: A corpus of
  {W}all {S}treet {J}ournal articles annotated with bridging links}.
\newblock In \emph{Proceedings of the Eleventh International Conference on
  Language Resources and Evaluation ({LREC} 2018)}, Miyazaki, Japan. European
  Language Resources Association (ELRA).

\bibitem[{R{\"o}ttger et~al.(2022)R{\"o}ttger, Vidgen, Hovy, and
  Pierrehumbert}]{rottger-etal-2022-two}
Paul R{\"o}ttger, Bertie Vidgen, Dirk Hovy, and Janet Pierrehumbert. 2022.
\newblock \href {https://doi.org/10.18653/v1/2022.naacl-main.13} {Two
  contrasting data annotation paradigms for subjective {NLP} tasks}.
\newblock In \emph{Proceedings of the 2022 Conference of the North American
  Chapter of the Association for Computational Linguistics: Human Language
  Technologies}, pages 175--190, Seattle, United States. Association for
  Computational Linguistics.

\bibitem[{Schweitzer et~al.(2018)Schweitzer, Eckart, G{\"a}rtner, Falenska,
  Riester, R{\"o}siger, Schweitzer, Stehwien, and
  Kuhn}]{schweitzer-etal-2018-german}
Katrin Schweitzer, Kerstin Eckart, Markus G{\"a}rtner, Agnieszka Falenska,
  Arndt Riester, Ina R{\"o}siger, Antje Schweitzer, Sabrina Stehwien, and Jonas
  Kuhn. 2018.
\newblock \href {https://aclanthology.org/L18-1457} {{G}erman radio interviews:
  The {GRAIN} release of the {SFB}732 silver standard collection}.
\newblock In \emph{Proceedings of the Eleventh International Conference on
  Language Resources and Evaluation ({LREC} 2018)}, Miyazaki, Japan. European
  Language Resources Association (ELRA).

\bibitem[{Uryupina et~al.(2019)Uryupina, Artstein, Bristot, Cavicchio, Delogu,
  Rodr{\'i}guez, and Poesio}]{Uryupina2019AnnotatingAB}
Olga Uryupina, Ron Artstein, Antonella Bristot, Federica Cavicchio, Francesca
  Delogu, Kepa~Joseba Rodr{\'i}guez, and Massimo Poesio. 2019.
\newblock \href {https://api.semanticscholar.org/CorpusID:164858637}
  {Annotating a broad range of anaphoric phenomena, in a variety of genres: the
  {ARRAU} corpus}.
\newblock \emph{Natural Language Engineering}, 26:95 -- 128.

\bibitem[{Waseem(2016)}]{waseem-2016-racist}
Zeerak Waseem. 2016.
\newblock \href {https://doi.org/10.18653/v1/W16-5618} {Are you a racist or am
  {I} seeing things? annotator influence on hate speech detection on
  {T}witter}.
\newblock In \emph{Proceedings of the First Workshop on {NLP} and Computational
  Social Science}, pages 138--142, Austin, Texas. Association for Computational
  Linguistics.

\bibitem[{Weischedel et~al.(2011)Weischedel, Pradhan, Ramshaw, Palmer, Xue,
  Marcus, Taylor, Greenberg, Hovy, Belvin et~al.}]{weischedel2011ontonotes}
Ralph Weischedel, Sameer Pradhan, Lance Ramshaw, Martha Palmer, Nianwen Xue,
  Mitchell Marcus, Ann Taylor, Craig Greenberg, Eduard Hovy, Robert Belvin,
  et~al. 2011.
\newblock Ontonotes release 4.0.
\newblock \emph{LDC2011T03, Philadelphia, Penn.: Linguistic Data Consortium},
  17.

\bibitem[{Yu et~al.(2022)Yu, Khosla, Manuvinakurike, Levin, Ng, Poesio, Strube,
  and Ros{\'e}}]{yu-etal-2022-codi}
Juntao Yu, Sopan Khosla, Ramesh Manuvinakurike, Lori Levin, Vincent Ng, Massimo
  Poesio, Michael Strube, and Carolyn Ros{\'e}. 2022.
\newblock \href {https://aclanthology.org/2022.codi-crac.1/} {The {CODI}-{CRAC}
  2022 shared task on anaphora, bridging, and discourse deixis in dialogue}.
\newblock In \emph{Proceedings of the CODI-CRAC 2022 Shared Task on Anaphora,
  Bridging, and Discourse Deixis in Dialogue}, pages 1--14, Gyeongju, Republic
  of Korea. Association for Computational Linguistics.

\bibitem[{Zeldes(2017)}]{Zeldes2017}
Amir Zeldes. 2017.
\newblock \href {https://doi.org/http://dx.doi.org/10.1007/s10579-016-9343-x}
  {The {GUM} corpus: Creating multilayer resources in the classroom}.
\newblock \emph{Language Resources and Evaluation}, 51(3):581--612.

\bibitem[{Zeldes(2022)}]{Zeldes2022}
Amir Zeldes. 2022.
\newblock \href {https://doi.org/10.5210/dad.2022.102} {Can we fix the scope
  for coreference? {P}roblems and solutions for benchmarks beyond {OntoNotes}}.
\newblock \emph{Dialogue \& Discourse}, 13(1):41--62.

\bibitem[{Zhang and Zeldes(2017)}]{Zhang2017GitDOXAL}
Shuo Zhang and Amir Zeldes. 2017.
\newblock \href {https://aaai.org/papers/619-flairs-2017-15451/} {{GitDOX}: A
  linked version controlled online {XML} editor for manuscript transcription}.
\newblock In \emph{Proceedings of the Thirtieth International Florida
  Artificial Intelligence Research Society Conference (FLAIRS 2017)}, pages
  619--623.

\end{thebibliography}

\appendix

\section{Subtypes in GUMBridge Test}
\label{sec:appendix_counts}

Table \ref{tab:adj_counts} shows the counts of the bridging subtypes in the adjudicated version of GUMBridge test v0.1.

\section{Subtypes by Genre in GUMBridge Test}
\label{sec:appendix_genre}

\begin{figure}
  \centering
  \includegraphics[width=1\linewidth]{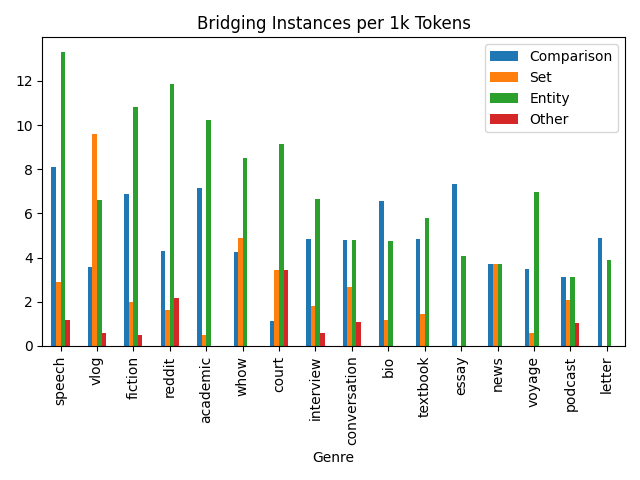}
  \caption{Counts of bridging relation types by genre in adjudicated GUMBridge data.}
  \label{fig:genre_bridgetype}
\end{figure}

Figure \ref{fig:genre_bridgetype} shows the number of bridging instances per 1k tokens of each bridging relation type (\textsc{comparison}, \textsc{set}, \textsc{entity}, and \textsc{other}) in each of the 16 genres in GUMBridge test (v0.1).

\begin{table}
\centering
\resizebox{0.55\columnwidth}{!}{%
\begin{tabular}{ll}
\hline
\textsc{\textbf{Comparison}} &  \\
\hspace{5mm}\textsc{relative} & 59 \\
\hspace{5mm}\textsc{time} & 27 \\
\hspace{5mm}\textsc{sense} & 45 \\
\hline
\hspace{5mm}Subtotal & 131 \\ \hline
\textsc{\textbf{Entity}} & \\
\hspace{5mm}\textsc{associative} & 124 \\
\hspace{5mm}\textsc{meronomy} & 37 \\
\hspace{5mm}\textsc{property} & 9 \\
\hspace{5mm}\textsc{resultative} & 21 \\
\hline
\hspace{5mm}Subtotal & 191 \\ \hline
\textsc{\textbf{Set}} & \\
\hspace{5mm}\textsc{member} & 31 \\
\hspace{5mm}\textsc{subset} & 14 \\
\hspace{5mm}\textsc{span-interval} & 18 \\
\hline
\hspace{5mm}Subtotal & 63 \\ \hline
\textsc{\textbf{Other}} & 16 \\ \hline
\textbf{Total} & 401 \\ \hline
\end{tabular}%
}
\caption{Counts of bridging subtypes in adjudicated GUMBridge data.}
\label{tab:adj_counts}
\end{table}

\section{Comparison with ARRAU Bridging Subtypes}
\label{sec:appendix_comp}

In order to allow for better comparison between the resources of GUMBridge and ARRAU, we include a brief comparison of how ARRAU's bridging subtypes\footnote{As the GUMBridge schema does not differentiate the relative roles of the anaphor and antecedent in the subtype relation, ARRAU's inverse subtypes map the same as their regular subtypes.} map onto the proposed schema for GUMBridge:

\paragraph{possession $\rightarrow$} Part-of relations that will mostly fall under \textsc{entity-meronomy} or \textsc{entity-property}. 

\paragraph{element-set $\rightarrow$} Maps to \textsc{set-member}.

\paragraph{subset-set $\rightarrow$} Maps to \textsc{set-subset}.

\paragraph{‘other’ anaphora $\rightarrow$} Maps to \textsc{comparison-relative}, which encompasses additional comparative markers not covered in ARRAU, including ordinals and comparative adjectives.

\paragraph{under-specified $\rightarrow$} \textsc{entity-associative} unless one of the other \textsc{entity} subtypes is a better fit based on the context. However, sense anaphora (\underline{green shirt} $\rightarrow$ \textbf{red one}) should be mapped to \textsc{comparative-sense}.

\end{document}